\documentclass{article}

\usepackage{PRIMEarxiv}

\usepackage[utf8]{inputenc} 
\usepackage[T1]{fontenc}    
\usepackage{hyperref}       
\usepackage{url}            
\usepackage{booktabs}       
\usepackage{amsfonts}       
\usepackage{nicefrac}       
\usepackage{microtype}      
\usepackage{lipsum}
\usepackage{fancyhdr}       
\usepackage{graphicx}       
\graphicspath{{media/}}     
\usepackage{amsmath} 
\usepackage{tikz}
\usepackage{subcaption}
\usetikzlibrary{arrows.meta}
\usepackage{array}
\title{ Predicting the Evolution of Interfaces with Fourier Neural Operators}

\author{
  Paolo Guida, \\
  Clean Energy Research Center (CERP) \\
  King Abdullah University of Science and Technology (KAUST) \\
  Thuwal 23955, Saudi Arabia\\
  \texttt{paolo.guida@kaust.edu.sa} \\
   \And
  William L. Roberts \\
  Clean Energy Research Center (CERP) \\
  King Abdullah University of Science and Technology (KAUST) \\
  Thuwal 23955, Saudi Arabia\\
  \texttt{william.roberts@kaust.edu.sa} \\
}

\begin{document}
\maketitle


\maketitle

\begin{abstract}
Recent progress in AI has established neural operators as powerful tools that can predict the evolution of partial differential equations, such as the Navier-Stokes equations. Some complex problems rely on sophisticated algorithms to deal with strong discontinuities in the computational domain. For example, liquid-vapour multiphase flows are a challenging problem in many configurations, particularly those involving large density gradients or phase change. The complexity mentioned above has not allowed for fine control of fast industrial processes or applications because computational fluid dynamics (CFD) models do not have a quick enough forecasting ability. This work demonstrates that the time scale of neural operators-based predictions is comparable to the time scale of multi-phase applications, thus proving they can be used to control processes that require fast response. Neural Operators can be trained using experimental data, simulations or a combination. In the following, neural operators were trained in volume of fluid simulations, and the resulting predictions showed very high accuracy, particularly in predicting the evolution of the liquid-vapour interface, one of the most critical tasks in a multi-phase process controller.
\end{abstract}

\section{Introduction}

Multiphase flow systems, such as liquid-vapour interactions, are ubiquitous in natural and industrial contexts \cite{ishii2010thermo, tryggvason2011direct}. Accurate prediction of interface dynamics in such systems is crucial for efficient design and control processes in energy production, chemical engineering, and environmental applications \cite{prosperetti2009computational, brackbill1992continuum}.

Traditional computational fluid dynamics (CFD) methods such as the volume of fluid (VoF) \cite{hirt1981volume, scardovelli1999direct} and front-tracking approaches \cite{tryggvason2001front} have been extensively developed to simulate multiphase flows with interface tracking. However, these methods often require fine grids and small time steps to resolve steep gradients and discontinuities, leading to high computational costs \cite{rudman1998volume, popinet2003gerris, roenby2017new}. This limits their applicability in real-time or embedded control scenarios where fast predictions are critical.

The recent rise of machine learning techniques tailored to scientific computing has opened new avenues for modelling complex systems of PDEs. Physics-Informed Neural Networks (PINNs) \cite{raissi2019physics} and operator-learning frameworks like DeepONets \cite{lu2021learning} and Neural Operators (NOs) \cite{azizzadenesheli2024neural} have shown exceptional promise in solving or approximating PDEs without requiring explicit discretization.

Among these, Fourier Neural Operators (FNOs) \cite{li2020fourier} stand out for their ability to efficiently learn mappings between function spaces and operate at arbitrary resolutions. Their spectral convolution layers leverage the fast Fourier transform to capture long-range dependencies, making them particularly effective for problems involving spatially extended dynamics \cite{li2020fourier, kovachki2021neural}. FNOs have already demonstrated strong performance in modelling turbulent flows \cite{thuerey2021pbdl}, weather forecasting \cite{pathak2022fourcastnet}, and reaction-diffusion systems \cite{brandstetter2022message}.

Despite these advances, applications of neural operators to multiphase flows remain limited. Multiphase systems pose additional challenges such as large density ratios, interface instability, and surface tension effects \cite{duraisamy2019turbulence,popinet2018numerical, vinuesa2022enhancing}. Recent works have begun addressing these using hybrid physics-based and data-driven approaches \cite{markidis2021ai4s, gupta2022towards}.

In this study, we investigate the potential of FNOs for real-time prediction of interface evolution in liquid-vapour systems. We trained a neural operator using VoF simulation data to learn the underlying physics. Our results show that FNOs replicate the interface dynamics accurately, enabling predictions in a fraction of the time needed for conventional CFD. 
This capability enables the application of numerical simulations in multiphase systems control that require extremely fast predictions.

\section*{Methodology}

We present a pipeline focused on constructing and training a data-driven model for predicting dynamic liquid-vapour multiphase systems in time. The prediction target is the volume fraction field $\alpha(x, y, t)$, which describes the interface between fluid phases and evolves due to complex flow dynamics. This task is particularly suited for neural operators, and in this study, we use the FNOs \cite{li2020fourier} to develop an efficient surrogate model.

\paragraph*{Volume of fluid method.}

We first briefly describe the solver used in this work, \texttt{compressibleInterIsoFoam} \cite{GAMET2020104722}, part of the OpenFOAM framework \cite{jasak2007openfoam}. The solver utilises the \textit{isoAdvector} method, based on the geometric volume-of-fluid (Vof) \cite{roenby2017new,scheufler2019accurate}. 
The modelling approach is based on the Eulerian Vof framework, initially proposed in \cite{hirt1981volume}. The dynamics of two-phase compressible fluids consisting of liquid and gas are described by unified velocity $\textbf{u}(\textbf{x},t)$, pressure $p(\textbf{x},t)$ and temperature $T(\textbf{x},t)$ fields.
The transport equation for the volume fraction $\alpha$ of a compressible fluid is expressed as:
\begin{equation}
\frac{\partial (\alpha\rho)}{\partial t}+\nabla \cdot (\alpha \rho \textbf{u})=0,
\label{volumeFraction}
\end{equation}
where $\rho$ is the density. 

In the context of a finite volume representation, at a given time, the variable $\alpha_l(t)$ determines the volume fraction of liquid in a given computational cell:

\begin{equation}
\alpha_l(t)\equiv\frac{1}{V_l}\int_{\Omega_l}{f(\textbf{x},t) dV}
\end{equation}
where $V_l$ is the volume of the $l$-th cell and $\Omega \in\mathbb{R}^{d}$ is the computational domain.  Note that the liquid volume fraction is a continuous variable, although it is defined as a cell-averaged quantity. The liquid volume fraction, therefore, takes the following values:

\begin{equation}
    \alpha(\textbf{x},t)=    
    \begin{cases}
      1 & \text{within the liquid}\\
      ]0,1[ & \text{at the interface}\\
      0 & \text{within the gas}
    \end{cases}  
\end{equation}

For a consistent indicial notation, $\alpha_1 = \alpha$ is redundantly defined as the liquid volume fraction while $\alpha_2 = 1-\alpha$ is the gas volume fraction. 
All local physical quantities are calculated as a linear combination of the properties of the two phases.  The physical properties experienced a discontinuity at the interface between liquid and gas.
The local quantity in the liquid and gas phases also depends, in general, on the composition of the mixture. The mixing rules used in this work are specified in the next section and in the results.
A general physical property $\xi$ is expressed as:
\begin{equation}
    \xi=\alpha_1\xi_1+\alpha_2\xi_2
    \label{properties}
\end{equation}
where subscripts 1 and 2 denote the liquid and gas phases, respectively. The mass and momentum conservation equations describe the bulk fluid motion 
\begin{equation}
    \frac{\partial\rho}{\partial t}+\nabla\cdot(\rho\textbf{u})=0 ,  
    \label{continuity}
\end{equation}
\begin{equation}
    \frac{\partial(\rho\textbf{u})}{\partial t}+\nabla\cdot(\rho\textbf{u}\textbf{u})=\nabla\cdot[\mu(\nabla\textbf{u}+\nabla\textbf u^T)]-\nabla p+\textbf{f}_s+\rho \textbf{g} \,,
    \label{momentum}
\end{equation}
where \textbf{g} represents the gravitational acceleration, $\mu$ is the dynamic viscosity, and $\textbf{f}_s$ is the surface tension force. The latter is generally modelled with the continuum method introduced by Brackbill et al.~\cite{BRACKBILL1992335}, 
\begin{equation}
    \textbf{f}_s=\sigma q\textbf{n} \,,
    \label{surfaceTensionForce}
\end{equation}
where $\sigma$ is the surface tension coefficient, \textbf{n} is the unit vector normal to the interface, and \textit{q} is the curvature. 
The present study adopts the geometric curvature reconstruction, proposed by \cite{CUMMINS2005425} and later adopted by other authors \cite{GAMET2020104722,guida2022computational}, in which the curvature is calculated by the reconstructed dist ance function (RDF), $\zeta$, as $q=-\nabla\cdot\left(\nabla\zeta/|\nabla\zeta|\right)$.

\paragraph{Neural operators.} Neural Operators (NOs) are a class of operator learning models that approximate mappings between infinite-dimensional function spaces \cite{li2020fourier}:
\begin{equation}
    \mathcal{G}: \mathcal{A}(\Omega;\mathbb{R}^m) \rightarrow \mathcal{U}(\Omega;\mathbb{R}^n),
\end{equation}
where both \( \mathcal{A}(\Omega;\mathbb{R}^m) \) and \( \mathcal{U}(\Omega;\mathbb{R}^n) \) are Banach spaces, and \( \mathcal{G} \) is a nonlinear operator that maps input functions \( a \in \mathcal{A} \) to output functions \( u \in \mathcal{U} \). Here,  \( \Omega\subset \mathbb{R}^d \) is a spatial domain with \( d \in \mathbb{N} \). Note that the input and output vector spaces \( \mathbb{R}^m \) and \( \mathbb{R}^n \) might have different dimensions. In this setting, \( a(x) \) may represent an input field such as a coefficient, boundary condition, or source term, \( u(x) \) is the output of the nonlinear operator applied to \( a \), typically the solution of a system of PDEs.

\paragraph{Fourier neural operators.} Unlike traditional architectures that use local convolutional filters, the Fourier Neural Operator (FNO) learns an approximation \( \mathcal{G}_\theta \approx \mathcal{G} \) by composing layers that act globally on functions across the entire domain \( \Omega \), making FNOs suitable for capturing long-range dependencies rather than only local information. The methodology proposed by Li et al. \cite{li2020fourier} begins by lifting the input function \( a(x) \) into a higher-dimensional representation:
\begin{equation}
    v_0(x) = P(a(x)),
\end{equation}
where \( a(x) \colon \Omega \rightarrow \mathbb{R}^m \) is the input function, and \( P: \mathbb{R}^m \rightarrow \mathbb{R}^{d_v} \) is a learned linear transformation. As anticipated, the purpose of \( P \) is to project the input into a (generally) higher-dimensional latent space \( \mathbb{R}^{d_v} \), enabling the neural operator to process richer data representations and capture more complex structures and dependencies. The output \( v_0(x) \in \mathbb{R}^{d_{v_0}}\) serves as the input to the subsequent operator layers. The operator layers are defined as:
\begin{equation}
    v_{l+1}(x) := \sigma\left( W v_l(x) + \left( \mathcal{K}(a; \phi)v_l \right)(x) \right), \quad \forall x \in D,
\end{equation}
where \( v_l(x) \) is the feature at location \( x \) at layer \( l \), \( W :\mathbb{R}^{d_v}\rightarrow\mathbb{R}^{d_v}\) is a learned local linear transformation, and \( \sigma \) is a nonlinear activation function. The nonlocal operator \( \mathcal{K}(a;\phi) \) is defined as:
\begin{equation}
    \left( \mathcal{K}(a; \phi)v_l \right)(x) := \int_\Omega \kappa(x, y, a(x), a(y); \phi) \, v_l(y) \, dy,
    \label{kernel_integral_operator}
\end{equation}
where \(\phi\) contains all weights and biases of the neural network that realises \(k:\mathbb{R}^{2d\times 2m }\rightarrow\mathbb{R}^{d_v \times d_v}\). In the context of FNOs, the kernel integral operator is efficiently approximated by a convolution in Fourier space:
\begin{equation}
    \left( \mathcal{K}(a;\phi)v_l \right)(x) = \mathcal{F}^{-1} \left( \mathcal{F}(\kappa_\phi) \cdot \mathcal{F}(v_l) \right)(x),
    \label{inverse_fourier_transform}
\end{equation}
where \( \mathcal{F} \) and \( \mathcal{F}^{-1} \) denote the Fourier transform and its inverse, respectively: 
\begin{equation}
    \left( \mathcal{F}v \right)_j(k) = \int_\Omega v_j(x) e^{-2\pi i \langle k,x\rangle} \, dx,
\end{equation}
and
\begin{equation}
    \left( \mathcal{F}^{-1}v \right)_j(x) = \int_\Omega v_j(k) e^{2\pi i \langle k,x\rangle} \, dk,
\end{equation}

here, \( i \) denotes the imaginary unit, and \( \hat{v} \) represents the Fourier coefficients $j=1,..., d_v$. To enhance translation invariance, the kernel’s dependence on the function values is removed, reducing it to depend only on the relative spatial displacement:
\begin{equation}
    \kappa(x, y, a(x), a(y)) = \kappa(x - y),
    \label{simplification}
\end{equation}
where \( \kappa: \mathbb{R}^d \to \mathbb{R}^{d_v \times d_v} \).
This assumption simplifies learning: the kernel depends only on \( x - y \) and can be efficiently approximated in Fourier space. Substituting Eq.~\eqref{simplification} into Eq.~\eqref{kernel_integral_operator}, the integral simplifies into a standard convolution:
\begin{equation}
    \left( \mathcal{K}(a;\phi)v_l \right)(x) = \int_\Omega \kappa(x - y) v_l(y) \, dy = (\kappa * v_l)(x),
\end{equation}
where \( * \) denotes convolution. Applying the convolution theorem, the kernel integral operator becomes:
\begin{equation}
    \left( \mathcal{K}_\phi v_l \right)(x) = \mathcal{F}^{-1} \left( R_\phi(k) \cdot \mathcal{F}(v_l)(k) \right)(x),
\end{equation}
where \( R_\phi(k) \in \mathbb{C}^{d_v \times d_v} \) is the learned Fourier-space kernel matrix. The tensor \( R_\phi \) is parametrised in \( \mathbb{C}^{k_{\max} \times d_v \times d_v} \), where \( k_{\max} \) is the maximum number of Fourier modes considered. Since \( \kappa \) is truncated to \( k_{\max} \) modes, \( \mathcal{F}(v_l) \) is similarly truncated. Identifying \( \mathcal{F}(v_l) \) with \( \hat{v_l} \), the Fourier layer operation becomes:
\begin{equation}
    (R \cdot \hat{v_l})_{k,h} = \sum_{j=1}^{d_v} R_{k,h,j} (\hat{v_l})_{k,j},
    \qquad
    k = 1,\dots,k_{\max}, \quad h = 1,\dots,d_v.
\end{equation}
The tensor \( R \) contains learnable parameters for the lowest frequencies. After applying the inverse Fourier transform, the update reads:
\begin{equation}
    v_{l+1}(x) = \sigma\left(W_lv_l(x)+\sum_{|k| < k_{\max}} R_l(k) \cdot \widehat{v_l}(k) e^{2\pi i \langle k, x\rangle} + b_l(x) \right),
\end{equation}
applied recursively as:
\begin{gather}
    v_1 = \sigma\left( W_0 v_0 + R_0 \cdot v_0 + b_0 \right), \\
    v_2 = \sigma\left( W_1 v_1 + R_1 \cdot v_1 + b_1 \right), \\
    \vdots \\
    v_L = \sigma\left( W_{L-1} v_{L-1} + R_{L-1} \cdot v_{L-1} + b_{L-1} \right).
\end{gather}
The final lifted representation \( v_L(x) \) is mapped to the output function via a projection:
\begin{equation}
    u(x) = Q(v_L(x)),
\end{equation}
where \( Q \) is a linear map. The model is trained by minimising the empirical loss:
\begin{equation}
    \mathcal{L}(\theta) = \frac{1}{N} \sum_{i=1}^N \| \mathcal{G}_\theta(a_i) - u_i \|_{L^2(\Omega)}^2,
\end{equation}

Each spectral layer aggregates global spatial information, making FNOs particularly efficient at modeling long-range dependencies in PDE solutions. The truncation to a limited number of Fourier modes introduces a spectral bias that promotes generalisation. The computational complexity per layer is:
\begin{equation}
    \mathcal{O}(N \log N + k_{max}N),
\end{equation}
where \( N \) is the number of spatial points and \( k_{max} \) the number of retained Fourier modes.

\section{Validation Metrics}

To evaluate the performance of the model, we employ four diagnostic metrics that collectively assess accuracy, variance capture, and stability:

\begin{itemize}
    \item \textbf{Mean Squared Error (MSE):} Quantifies the average squared difference between predicted and true values:
    \begin{equation}
        \text{MSE}(t) = \frac{1}{N} \sum_{i=1}^{N} \left( \hat{\alpha}_{t,i} - \alpha_{t,i} \right)^2,
    \end{equation}
    where \( \hat{\alpha}_{t,i} \) and \( \alpha_{t,i} \) denote the predicted and true values at time \( t \) and spatial index \( i \).

    \item \textbf{Coefficient of Determination (\(R^2\)):} Measures the proportion of variance in the true data captured by the model:
    \begin{equation}
        R^2(t) = 1 - \frac{\sum_{i=1}^{N} \left( \hat{\alpha}_{t,i} - \alpha_{t,i} \right)^2}{\sum_{i=1}^{N} \left( \alpha_{t,i} - \bar{\alpha}_t \right)^2},
    \end{equation}
    where \( \bar{\alpha}_t \) is the spatial mean of the true field at time \( t \).

    \item \textbf{L2 Norm Error:} Captures the global deviation between the predicted and true fields:
    \begin{equation}
        E_t = \left\| \hat{\alpha}_t - \alpha_t \right\|_2,
    \end{equation}
    where \( \| \cdot \|_2 \) denotes the standard Euclidean norm over all spatial points.

    \item \textbf{Relative Error (RE):} Normalizes the L2 error by the magnitude of the true field:
    \begin{equation}
        \text{RE}(t) = \frac{\left\| \hat{\alpha}_t - \alpha_t \right\|_2}{\left\| \alpha_t \right\|_2}.
    \end{equation}
\end{itemize}

These metrics provide a comprehensive assessment framework for validating operator-based surrogate models, particularly in complex settings such as multiphase flow simulations. They characterise the model’s predictive accuracy, variance-capturing ability, and temporal robustness.

\paragraph{Network architecture.}
The surrogate model architecture comprises five stacked two-dimensional Fourier layers, each followed by a ReLU activation function and layer normalization.
The input to the network comprises two channels: the initial SDF field and a scalar-valued time channel. The output is a single-channel SDF prediction. We fix the number of Fourier modes per spatial dimension to $k_{\text{max}} = 20$, and use $d_v = 96$ hidden channels uniformly across all layers.

\paragraph{Optimization.}
Training is performed using the AdamW optimizer \cite{kingma2014adam,zhang2018improved} for 100 epochs with an initial learning rate of $5 \times 10^{-4}$ and a weight decay of $10^{-4}$.
We minimize the $L^2$ operator norm employing a batch size of 32. For Case~1, the dataset is split 60\% for training and 40\% for validation to evaluate temporal extrapolation. For Case~2, where generalization across initial conditions is the focus, a 90\%/10\% train-validation split is used.
All data are randomly shuffled and batched at each epoch.

\paragraph{Hardware.}
All experiments are conducted on a single \textsc{Nvidia} Quadro RTX 4000 GPU with 8\,GB of memory.
Training on a $N = 900$ base simulation dataset (yielding 1800 SDF samples) takes approximately $1200 s$.

\section{Results}
The proposed pipeline is demonstrated on two distinct benchmark scenarios. In the first case, we consider the gravitational collapse of a liquid column initially shaped as a rectangular body. The model is trained on the evolution of the interface over the first 60\% of the total simulation duration, and tasked with predicting the remaining 40\% purely from the learned operator dynamics.
This setup evaluates the model’s ability to extrapolate spatio-temporal behavior beyond the training horizon.
\begin{figure}[ht!]
    \centering
    \includegraphics[width=0.625\linewidth]{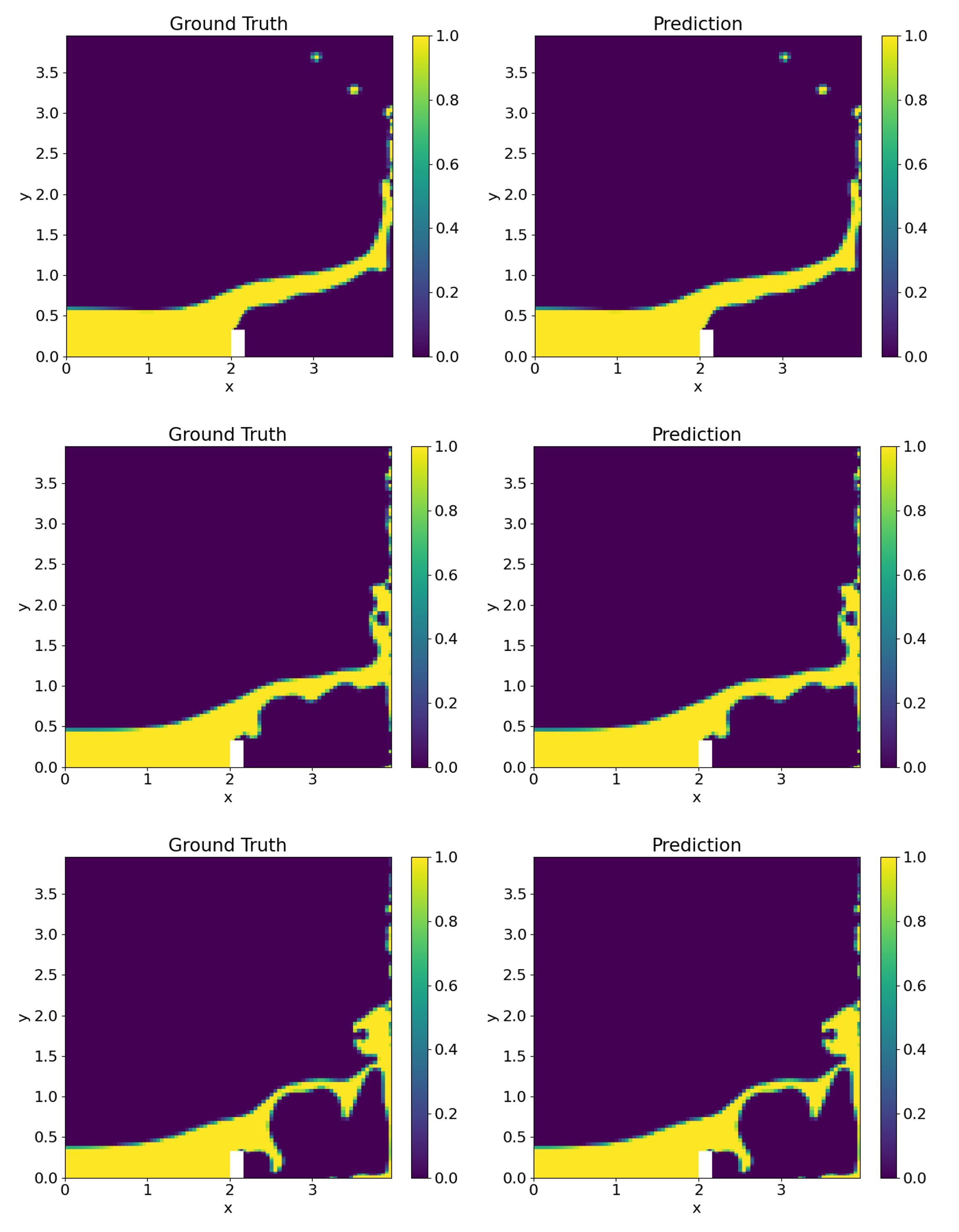}
 da   \caption{Time evolution of the liquid interface in the dam-break scenario. The figure shows three predicted time-steps comparing ground truth with the outcome of the FNO. The model is trained on the first 60\% of the time sequence and evaluated on the remaining 40\%. The mesh resolution is $164\times164$ }
    \label{fig:evolution-dambreak}
\end{figure}

The second case involves a more complex dataset of randomised fluid configurations evolving under gravity and surface tension. Each training sample includes two temporally spaced snapshots, and the objective is to learn the transition dynamics across diverse initial conditions.
This scenario assesses the model's generalisation ability across varied flow morphologies and temporal resolutions.
\paragraph{Case 1: forecasting.}
This study applies the \emph{Fourier Neural Operator} (FNO) to a two-phase dam-break simulation.
The dataset comprises $T$ simulations performed using \texttt{compressibleInterIsoFoam}, a solver included in the OpenFOAM-2406 framework~\cite{jasak2007openfoam}. 
The primary variable of interest for each simulation is the liquid volume fraction, denoted by $\alpha(x, y, t)$, which encodes the liquid fraction at each finite volume in the domain.
To exploit the FNO's strength in learning continuous fields, we transform $\alpha$ into a reconstructed distance function (RDF), which provides a smooth approximation of the signed distance from the interface.
The transformation is achieved using a regularised inverse Heaviside mapping, based on the standard volume-of-fluid reconstruction formulation:
\begin{equation}
    \zeta(x, y, t) = \epsilon \cdot \tanh^{-1}\left(1 - 2\alpha(x, y, t)\right),
    \label{eq:alpha_to_rdf}
\end{equation}
where $\zeta$ denotes the RDF, and $\epsilon$ is a tunable smoothing parameter that controls the sharpness of the interface.
This preprocessing step leverages the smoothness of $\phi$ to enhance the FNO’s ability to learn spatial correlations and interface dynamics, making it especially well-suited for time-evolving, multiphase systems.

\begin{figure*}[ht]
    \centering
    \begin{subfigure}[b]{0.48\linewidth}
        \centering
        \includegraphics[width=\linewidth]{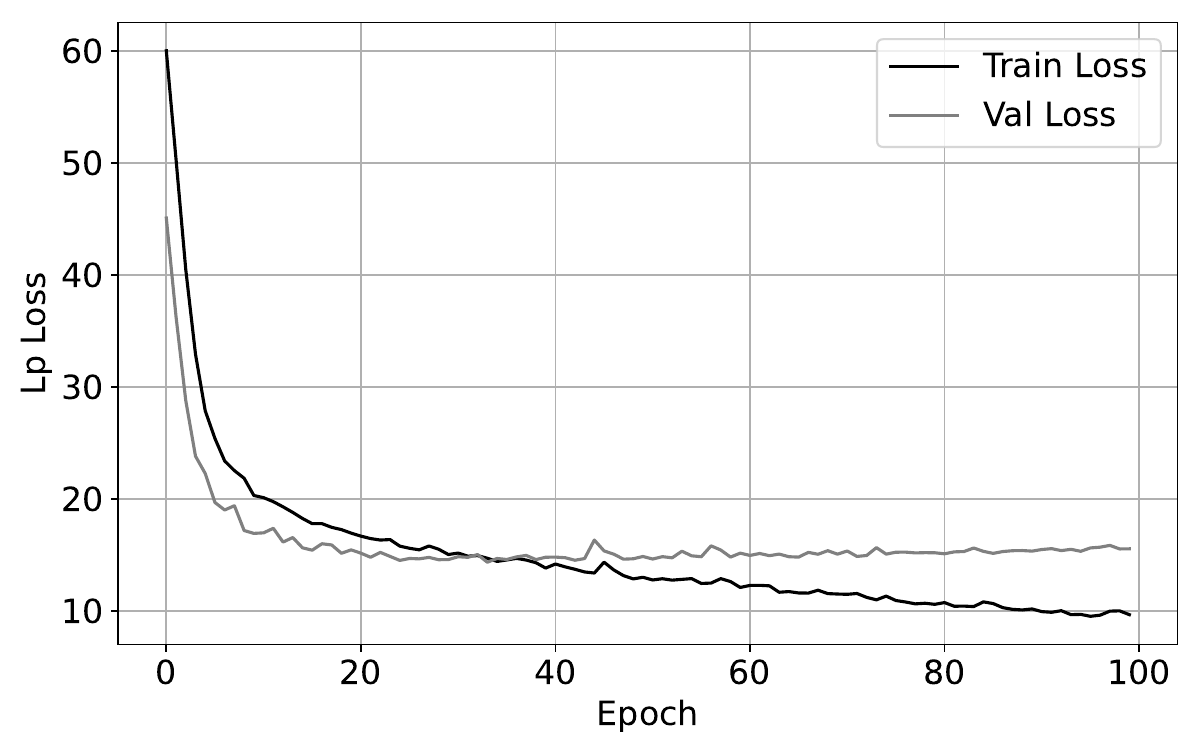}
        \caption{$L^p$ loss over training epochs for the dam-break simulation.}
        \label{fig:lp-loss}
    \end{subfigure}
    \hfill
    \begin{subfigure}[b]{0.48\linewidth}
        \centering
        \includegraphics[width=\linewidth]{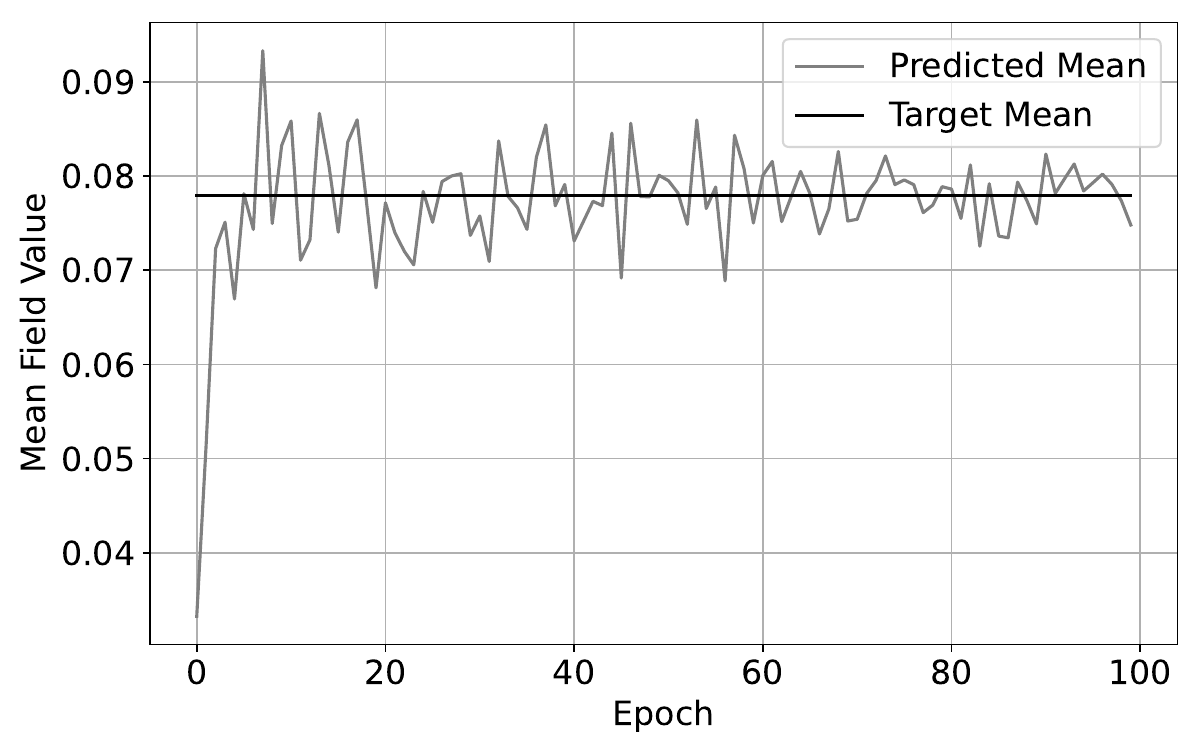}
        \caption{Predicted vs. ground-truth volume fraction.}
        \label{fig:mean-evolution}
    \end{subfigure}

    \vspace{1em}
    \begin{subfigure}[b]{0.98\linewidth}
        \centering
        \includegraphics[width=\linewidth]{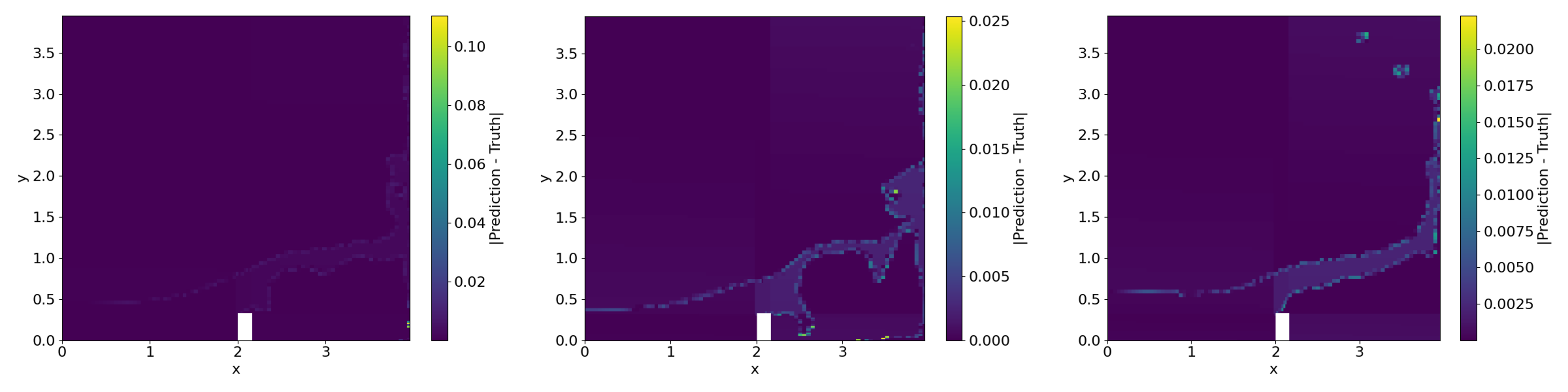}
        \caption{Spatial error distribution in three predicted snapshots from the test set.}
        \label{fig:error-snapshots}
    \end{subfigure}

    \caption{
    \textbf{Training diagnostics and spatial accuracy of the Fourier Neural Operator on the dam-break benchmark.}
    (\subref{fig:lp-loss}) The evolution of the $L^p$ loss over 100 training epochs indicates consistent convergence of the surrogate model.
    (\subref{fig:mean-evolution}) Compares the predicted and ground-truth spatial means of the signed-distance field (SDF) across training epochs, showing strong agreement and low variance.
    (\subref{fig:error-snapshots}) Visualizes the spatial distribution of absolute error for three representative frames sampled from the test set, illustrating the model's accuracy in resolving complex interface features and localized discrepancies.
    These diagnostics highlight the model’s robust training dynamics and generalization performance across unseen dam-break scenarios.
    }
    \label{fig:training-and-eval}
\end{figure*}
\begin{figure*}[ht]
    \centering
    \begin{subfigure}[b]{0.48\linewidth}
        \centering
        \includegraphics[width=\linewidth]{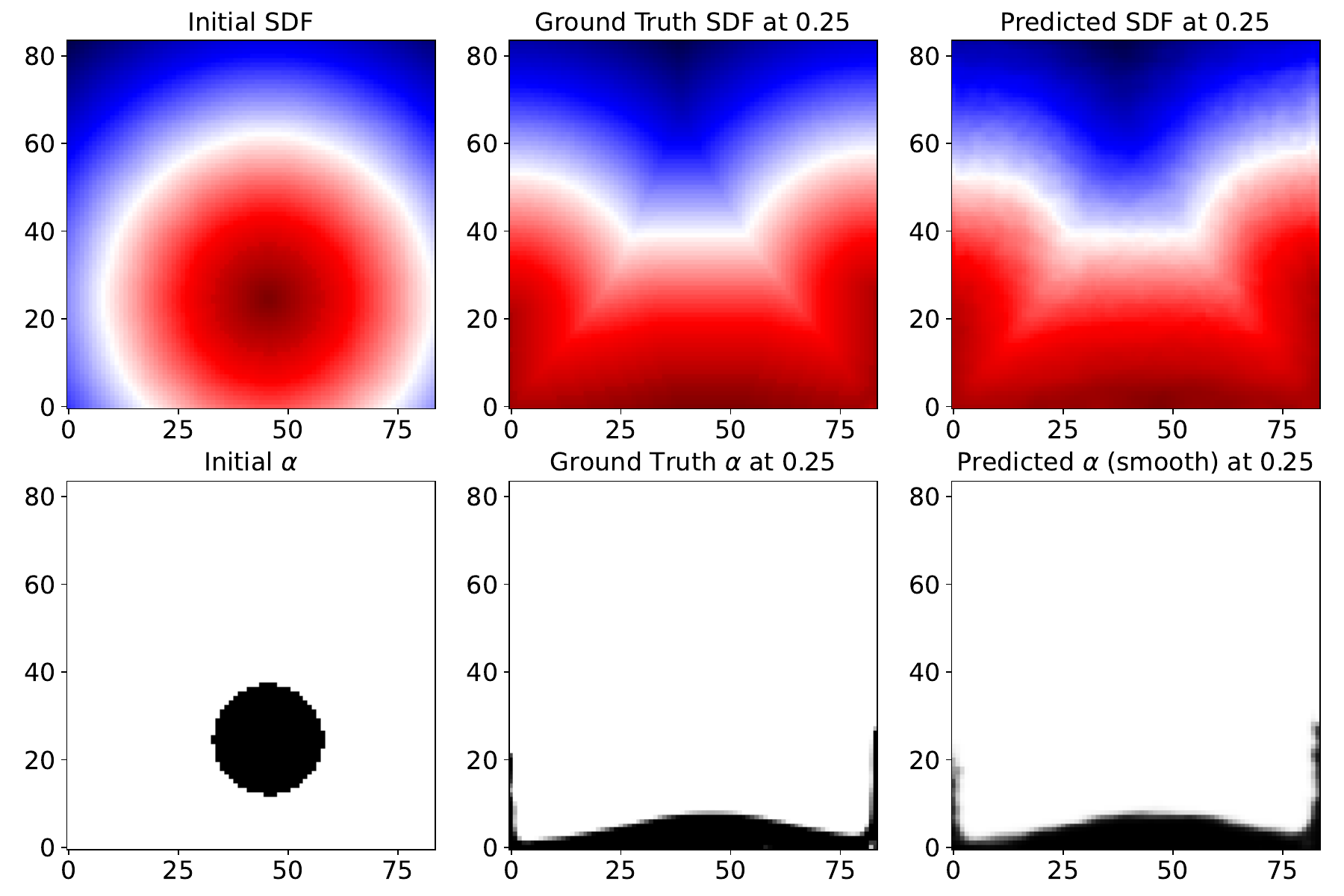}
        \caption{Predictions at $t = 0.25\,\mathrm{s}$}
        \label{fig:validation-025}
    \end{subfigure}
    \hfill
    \begin{subfigure}[b]{0.48\linewidth}
        \centering
        \includegraphics[width=\linewidth]{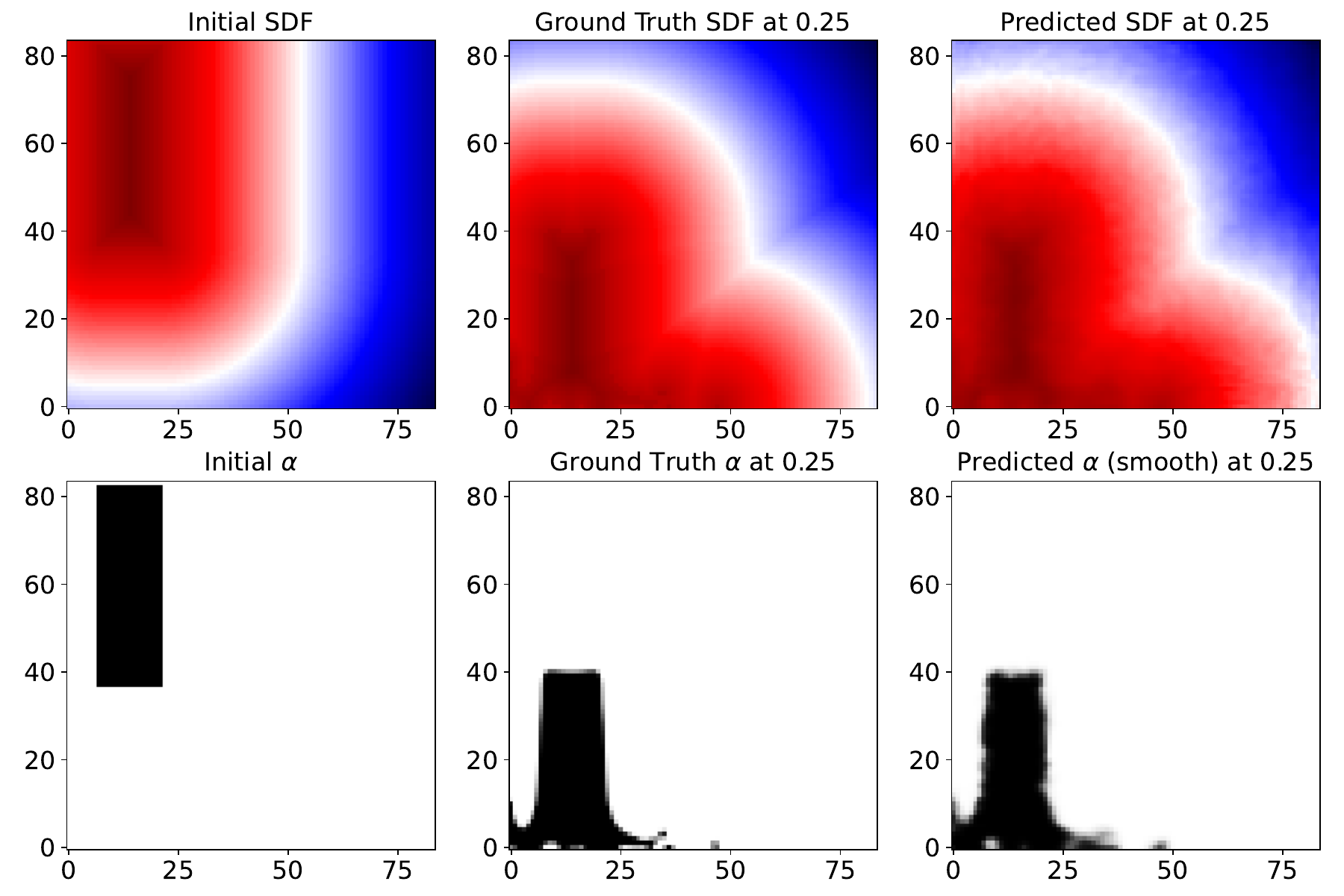}
        \caption{Predictions at $t = 0.50\,\mathrm{s}$}
        \label{fig:validation-050}
    \end{subfigure}
    
    \caption{
    \textbf{Comparison of predicted and ground-truth results at intermediate and final time steps.}
    Each subfigure presents a visual comparison between the \emph{initial input}, the \emph{ground-truth} at a given time, and the \emph{predicted} field from the FNO model.
    The top row of each panel shows the signed-distance fields (RDF): (left) the initial RDF at $t=0$, (centre) the ground truth at the evaluated time, and (right) the FNO prediction.
    The bottom row shows the corresponding binary volume fractions reconstructed from each RDF.
    (\subref{fig:validation-025}) Evaluates the model’s prediction at an intermediate time step ($t=0.25\,\mathrm{s}$).
    (\subref{fig:validation-050}) Shows predictions at the final time step ($t=0.50\,\mathrm{s}$), highlighting the model’s ability to extrapolate interface dynamics from initial conditions.
    The FNO effectively captures the evolving interface's global structure and fine-scale morphology in both cases.
    }
    \label{fig:rdf-validation-cluster}
\end{figure*}
\begin{table}[h!]
\centering
\renewcommand{\arraystretch}{1.5}
\caption{FNO Model Configuration}
\label{tab:fno_config}
\begin{tabular}{|>{\centering\arraybackslash}m{6cm}|>{\centering\arraybackslash}m{4cm}|}
\hline
\textbf{Parameter} & \textbf{Value} \\
\hline
Input Channels ($C_{in}$) & 2 \\
Output Channels ($C_{out}$) & 1 \\
Hidden Channels ($C_{hid}$) & 96 \\
Fourier Modes ($n_{modes}$) & (20, 20) \\
Number of Layers ($n_{layers}$) & 5 \\
\hline
\end{tabular}
\end{table}

\vspace{0.5cm}

\begin{table}[h!]
\centering
\renewcommand{\arraystretch}{1.5}
\caption{Validation Performance Metrics}
\label{tab:fno_metrics}
\begin{tabular}{|>{\centering\arraybackslash}m{6cm}|>{\centering\arraybackslash}m{4cm}|}
\hline
\textbf{Metric} & \textbf{Value} \\
\hline
Mean Squared Error (MSE) & 9.72\\
Mean Absolute Error (MAE) & 1.69 \\
$R^2$ Score & 0.95 \\
\hline
\end{tabular}
\end{table}
\paragraph{Case 2: learning dynamics.}
This test aims to learn the dynamics of a freely falling liquid blob in a two-dimensional dam-break scenario.
To this end, we created a dataset of 1\,000 random initial conditions, each generated by seeding distinct volumetric liquid regions into a fixed-size 2D domain.
The domain is discretized into $84 \times 84$ cells and spans a vertical slice of a fluid reservoir with free-surface evolution governed by the two-phase Navier-Stokes equations.

The volume-fraction field ($\alpha$) was recorded at three time instances: $t = 0.00\,\mathrm{s}$, $t = 0.25\,\mathrm{s}$, and $t = 0.50\,\mathrm{s}$.
All $\alpha$ fields were converted to signed-distance functions (SDFs) using the method mentioned above before training to facilitate gradient-based learning better.
For each simulation, we formed two labeled pairs: \{initial SDF, $t=0.25$\} and \{initial SDF, $t=0.50$\}, thereby yielding $2N = 2\,000$ samples for the FNO.
These were encoded as four-dimensional tensors with shape $[N, 2, 84, 84]$, where the second channel contains a scalar time field broadcast across the spatial grid.

We trained an FNO consisting of five spectral convolution layers with $(k_x, k_y) = (20, 20)$ Fourier modes, $d_v=96$ hidden channels, and ReLU activations.
Training was performed over 100 epochs using the AdamW optimizer \cite{kingma2014adam} and a learning rate of $5 \times 10^{-4}$, with 90\% of the data used for training and 10\% held out for validation.

\paragraph{Performance validation.}
\autoref{tab:fno_metrics} reports the performance of the FNO on the validation set.
The Fourier Neural Operator (FNO) model demonstrated strong predictive accuracy, achieving a mean squared error (MSE) of $9.72$ and a mean absolute error (MAE) of $1.69$ on the validation set when reconstructing signed-distance fields (SDFs) at future time steps.
The coefficient of determination ($R^2$ score) reached $0.95$, indicating excellent agreement with the ground truth and effective generalization to unseen spatio-temporal inputs.
These results confirm the model's capacity to interpolate complex two-phase interface dynamics despite the inherent nonlinearity and discontinuities typical of such flows.

\begin{figure}[!ht]
    \centering
    \includegraphics[width=0.65\linewidth]{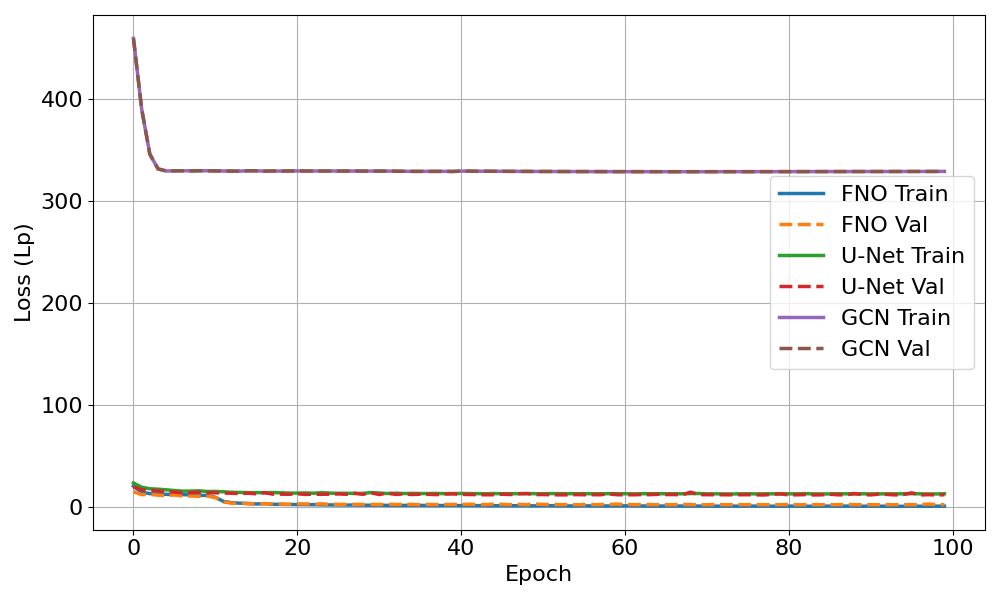}
    \caption{
    \textbf{Comparison of $L^p$ training and validation loss across architectures.}
    The FNO exhibits superior convergence behavior and consistently achieves the lowest validation loss, underscoring its ability to learn continuous operator mappings in complex flow regimes.
    While the U-Net performs well early in training, it levels off with a higher validation loss and greater variance.
    Although more efficient than FNO in training time, the GCN model fails to capture the full spatial-temporal dynamics and saturates at a significantly higher loss.
    The latter highlights the trade-off between accuracy and training cost: FNO (1257 s) vs. GCN (470 s) vs. U-Net (190 s).
    }
    \label{fig:model-loss-comparison}
\end{figure}

To contextualise it, we compared FNO's performance against two alternative neural architectures: a U-Net and a Graph Convolutional Network (GCN).
Figure~\ref{fig:model-loss-comparison} presents the training and validation $L^p$ loss over 100 epochs for each method, alongside their computational cost.
The training times recorded were 1257 seconds for FNO, 190 seconds for U-Net, and 470 seconds for GCN, reflecting a trade-off between model expressiveness and efficiency.

\section{Conclusions}
The pipeline developed in this work demonstrates how surrogate modelling based on Fourier Neural Operators (FNOs) enables efficient and accurate prediction of multiphase interface dynamics.
We achieve rapid forecasting capabilities for real-time control and scientific exploration by combining high-fidelity simulation data with operator learning.

We evaluate this approach on two representative applications. In the first, the FNO is trained on the early-time evolution of a dam-break scenario and is tasked with forecasting the subsequent interface dynamics, including impact events and nonlinear deformation.
The model successfully captures the underlying physics and extrapolates the evolution beyond the training window.

The second application focuses on learning the dynamics of freely falling liquid blobs in a 2D domain.
Here, the model is trained on a limited set of 900 randomly generated simulations, each represented by only two time steps.
Despite this sparse supervision, the FNO generalizes well across previously unseen configurations, accurately predicting the interface progression across diverse initial conditions. It is important to mention that the inference time for the models described in this work ranged from 1-8 ms underscoring the potential of FNOs to enable high-fidelity, low-latency control of complex multiphase systems, with significant implications for digital twin technologies and real-time monitoring in fluid mechanics applications. The next step will be to extend this approach to 3D systems and evaluate the ability of this framework to predict phase change.

\section*{List of Symbols}

\begin{itemize}
  \item $ \alpha $ — Volume fraction of the liquid phase
  \item $ \alpha_l(t) $ — Volume fraction of liquid in the $l$-th computational cell at time $t$
  \item $ \alpha_1 = \alpha $ — Liquid volume fraction
  \item $ \alpha_2 = 1 - \alpha $ — Gas volume fraction
  \item $ \rho $ — Fluid density
  \item $ \textbf{u}(\textbf{x}, t) $ — Velocity field
  \item $ p(\textbf{x}, t) $ — Pressure field
  \item $ T(\textbf{x}, t) $ — Temperature field
  \item $ \Omega \subset \mathbb{R}^d $ — Spatial computational domain
  \item $ V_l $ — Volume of the $l$-th finite volume cell
  \item $ \xi $ — Generic physical property (e.g., density, viscosity)
  \item $ \xi_1, \xi_2 $ — Property value in liquid and gas phases, respectively
  \item $ \textbf{g} $ — Gravitational acceleration vector
  \item $ \mu $ — Dynamic viscosity
  \item $ \textbf{f}_s $ — Surface tension force
  \item $ \sigma $ — Surface tension coefficient
  \item $ \textbf{n} $ — Unit normal vector to the interface
  \item $ q $ — Interface curvature
  \item $ \zeta $ — Reconstructed Distance Function (RDF)
  \item $ \mathcal{G} $ — True nonlinear operator mapping input to output
  \item $ \mathcal{G}_\theta $ — Neural operator approximation of $ \mathcal{G} $
  \item $ \mathcal{A}(\Omega; \mathbb{R}^m) $ — Input function space
  \item $ \mathcal{U}(\Omega; \mathbb{R}^n) $ — Output function space
  \item $ a(x) $ — Input function
  \item $ u(x) $ — Output function (e.g., PDE solution)
  \item $ v_l(x) $ — Intermediate feature at layer $l$
  \item $ d_v $ — Latent feature dimensionality
  \item $ P $ — Lifting operator: $ \mathbb{R}^m \to \mathbb{R}^{d_v} $
  \item $ Q $ — Projection operator: $ \mathbb{R}^{d_v} \to \mathbb{R}^n $
  \item $ W $ — Local linear transformation
  \item $ R_l(k) $ — Spectral kernel at layer $l$ and frequency $k$
  \item $ R_\phi(k) $ — Parametrized spectral kernel in Fourier space
  \item $ k_{\max} $ — Maximum number of Fourier modes retained
  \item $ \widehat{v}_l(k) $ — Fourier coefficients of $v_l(x)$
  \item $ \mathcal{F}, \mathcal{F}^{-1} $ — Fourier transform and inverse
  \item $ \kappa(x - y) $ — Convolution kernel in physical space
  \item $ \sigma $ — Nonlinear activation function 
  \item $ \hat{\alpha}_{t, i} $ — Predicted value of $\alpha$ at time $t$, index $i$
  \item $ \alpha_{t, i} $ — True value of $\alpha$ at time $t$, index $i$
  \item $ \bar{\alpha}_t $ — Mean of ground-truth $\alpha$ at time $t$
  \item $ N $ — Number of spatial points
  \item $ \text{MSE}(t) $ — Mean Squared Error at time $t$
  \item $ R^2(t) $ — Coefficient of determination at time $t$
  \item $ E_t $ — $L^2$ norm of prediction error at time $t$
  \item $ \text{RE}(t) $ — Relative error at time $t$
\end{itemize}

\section*{Acknowledgements}
The authors sincerely thank the open-source communities that made this work possible. In particular, we acknowledge the developers and contributors of the \textbf{Python} ecosystem, and \textbf{OpenFOAM} community. Special recognition is given to the developers of \texttt{compressibleInterIsoFoam} and related two-phase solvers. This research was supported by \textbf{King Abdullah University of Science and Technology (KAUST)}.

\bibliographystyle{unsrt}  
\bibliography{references}

\end{document}